\documentclass[letterpaper, 10 pt, conference]{ieeeconf}  

\IEEEoverridecommandlockouts                              

\usepackage{times}
\usepackage[english]{babel}
\usepackage[utf8]{inputenc}
\usepackage[T1]{fontenc}
\usepackage{fancyhdr}
\usepackage{verbatim}
\usepackage{layout}
\usepackage{float}
\usepackage[stable]{footmisc}
\usepackage{acronym}
\usepackage{amssymb}
\usepackage{amsmath}
\usepackage{graphicx}
\usepackage{url}
\usepackage{hyperref}
\usepackage{epstopdf}
\usepackage{epsfig}
\usepackage{listings}
\usepackage{setspace}
\usepackage[usenames]{color}
\usepackage{rotating}
\usepackage{algorithm}
\usepackage{algorithmic}
\usepackage{subfigure}
\usepackage{tikz}
\usetikzlibrary{calc, shapes, backgrounds}
\usepackage{pgfplots}
\usepackage{adjustbox}
\usepackage{comment}

\acrodef{emg}[EMG]{Electromyography}
\acrodef{semg}[sEMG]{Surface Electromyography}
\acrodef{mog}[MoG]{Mixture of Gaussians}
\acrodef{gmm}[GMM]{Gaussian Mixture Model}
\acrodef{gmr}[GMR]{Gaussian Mixture Regression}
\acrodef{nn}[NN]{Neural Networks}
\acrodef{fnn}[FNN]{Feedforward Neural Network}
\acrodef{eeg}[EEG]{Electroencephalogram}
\acrodef{nmse}[NMSE]{Normalized Mean Square Error}
\acrodef{mse}[MSE]{Mean Square Error}
\acrodef{rlfd}[RLfD]{Robot Learning from Demonstration}
\acrodef{em}[EM]{Expectation-Maximization}
\acrodef{bic}[BIC]{Bayesian Information Criterion}
\acrodef{aic}[AIC]{Akaike Information Criterion}
\acrodef{mdl}[MDL]{Minimum Description Length}
\acrodef{mml}[MML]{Minimum Message Length}
\acrodef{com}[CoM]{Center of Mass}
\acrodef{mav}[MAV]{Mean Average Value}
\acrodef{rms}[RMS]{Rooted Mean Square}
\acrodef{var}[VAR]{Standard Deviation}
\acrodef{dof}[DoF]{Degree of Freedom}
\acrodef{dofs}[DoFs]{Degrees of Freedom}
\acrodef{ft}[Fourier Transform]{Fourier Transform}
\acrodef{wt}[Wavelet Transform]{Wavelet Transform}
\acrodef{dwt}[DWT]{Discrete Wavelet Transform}
\acrodef{cwt}[CWT]{Continuous Wavelet Transform}
\acrodef{ros}[ROS]{Robot Operating System}
\acrodef{iav}[IAV]{Integral Absolute Value}
\acrodef{mpf}[MPF]{Mean Power Frequency}
\acrodef{hmm}[HMM]{Hidden Markov Model}
\acrodef{hmmod}[Hill Muscle Model]{Hill Muscle Model}
\acrodef{hpr}[HPR]{Hierarchical Projected Regression}
\acrodef{gof}[GoF]{Goodness of Fit}
\acrodef{hmi}[HMI]{Human-Machine Interfaces}
\acrodef{cca}[CCA]{Canonical Correlation Analysis}
\acrodef{imu}[IMU]{Inertial Measurement Unit}
\acrodef{acc}[ACC]{accelerometer}
\acrodef{pdf}[PDF]{Probability Density Function}
\acrodef{lda}[LDA]{Linear Discriminant Analysis}
\acrodef{svm}[SVM]{Support Vector Machine}
\acrodef{pca}[PCA]{Principal Components Analysis}
\acrodef{mocap}[MoCap]{Motion Capture}
\acrodef{aal}[AAL]{Active and Assisted Living}
\acrodef{fsm}[FSM]{Finite State Machine}
\acrodef{nmf}[NMF]{Non-negative Matrix Factorization}
\acrodef{vaf}[VAF]{Variance-Accounted-For}
\acrodef{fda}[FDA]{Fisher's Discriminant Analysis}
\newcommand{\fig}{Figure~}
\newcommand{\tab}{Table~}
\newcommand{\sect}{Section~}

\newcommand{\eq}{Equation~}

\title{\LARGE \bf
Fast human motion prediction for human-robot collaboration
with wearable interfaces}

\author{Stefano Tortora$^{1}$$^{\ddagger}$ and Stefano Michieletto$^{1}$$^{\ddagger}$, Francesca Stival$^{1}$ and Emanuele Menegatti$^{1}$
\thanks{$^{1}$Stefano Tortora, Stefano Michieletto, Francesca Stival and Emanuele Menegatti are with the Intelligent Autonomous Systems Lab (IAS-Lab), Department of Information Engineering (DEI), University of Padova {\tt\small {tortora, michieletto, stivalfr, emg}@dei.unipd.it}}%
\thanks{$\ddagger$ These authors contributed equally to this work.}
}

\begin{document}

\maketitle
\thispagestyle{empty}
\pagestyle{empty}

\begin{abstract}
In this paper, we aim at improving human motion prediction during human-robot collaboration in industrial facilities by exploiting contributions from both physical and physiological signals. 
Improved human-machine collaboration could prove useful in several areas, while it is crucial for interacting robots to understand human movement as soon as possible to avoid accidents and injuries.
In this perspective, we propose a novel human-robot interface capable to anticipate the user intention while performing reaching movements on a working bench in order to plan the action of a collaborative robot.
The proposed interface can find many applications in the Industry 4.0 framework, where autonomous and collaborative robots will be an essential part of innovative facilities.
A \textit{motion intention prediction} and a \textit{motion direction prediction} levels have been developed to improve detection speed and accuracy.
A Gaussian Mixture Model~(GMM) has been trained with IMU and EMG data following an evidence accumulation approach to predict reaching direction. Novel dynamic stopping criteria have been proposed to flexibly adjust the trade-off between early anticipation and accuracy according to the application. 
The output of the two predictors has been used as external inputs to a Finite State Machine~(FSM) to control the behaviour of a physical robot according to user's action or inaction.
Results show that our system outperforms previous methods, achieving a real-time classification accuracy of $94.3\pm2.9\%$ after $160.0msec\pm80.0msec$ from movement onset.
\end{abstract}

\acresetall
\section{Introduction}
\label{sec:intro}
The interaction between men and machines is a fundamental part in Industry 4.0~\cite{lasi2014industry}.
The availability of affordable and reliable collaborative robots opens new and interesting perspectives. Human and robot can work together in the same area, or even on the same product. 
This requires that the robot reliably predicts human motion in order to properly adjust its 
trajectory. The prediction should be performed in the early stages of the motion to ensure the collaboration to be safe.
\begin{figure}
\centering
\includegraphics[width=\columnwidth]{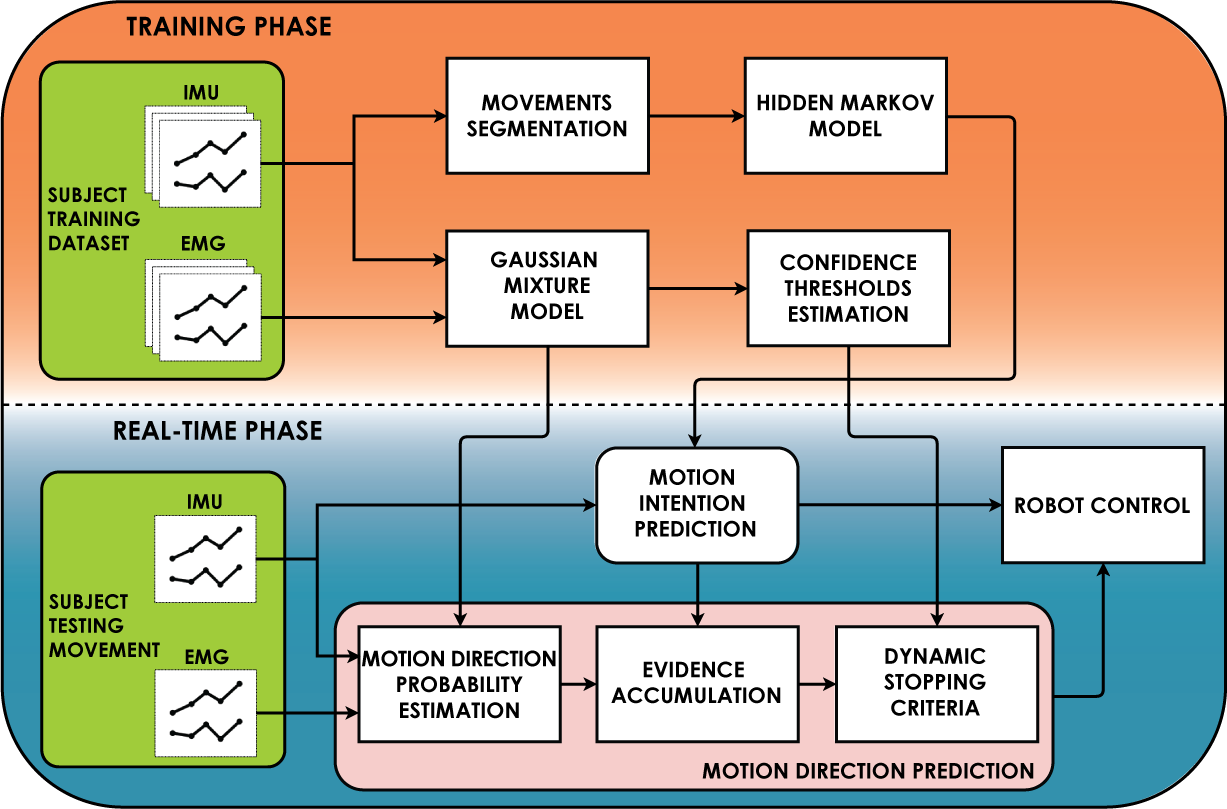}
  \caption{Human-robot interface framework for prediction of human motion using \acs{emg} and \acs{imu} information. Prediction models' parameters are estimated during the training phase from subject's data. The intention of performing a movement is detected and used to start the evidence accumulation of the movement direction prediction from a mixture model. Once the criteria of stopping are met, the prediction is sent to the robot controller.}
\label{fig:diagram}
\end{figure}
Many works focused on anticipating human intention to select the subsequent robot action~\cite{hawkins2013probabilistic}\cite{hawkins2014anticipating}\cite{hoffman2007cost}. These works did not consider kinematic information and require the system to know the possible human tasks sequences beforehand. Moreover, they do not directly tackle the problem of 
predicting human movements as soon as they begin.

In this paper, we present an approach exploiting \ac{imu} and \ac{emg} data to predict in real-time the target of a human while performing a reaching motion. The proposed system aims to (i)~detect when the subject is moving, (ii)~anticipate the identification of the movement direction, and (iii)~maximize the trade-off between prediction accuracy and early-anticipation, while being robust to possible misclassifications. An overview of the framework, implemented under the \ac{ros} middleware, is shown in \fig\ref{fig:diagram}.
In a preliminary work~\cite{tortora2018synergy}, we developed a predictive system for healthcare based on \ac{gmm}. An evidence accumulation framework selected the target starting by rich \ac{emg} signals. We have shown this approach to rapidly detect reaching direction close to movement onset, also in presence of pathological movements~\cite{tortorasynergy}. Now, we aim at applying such methods to human-robot collaboration by pairing \ac{imu} information to \ac{emg} data. \ac{imu} information has been used also to train a \ac{hmm} that predicts the beginning of a reaching movement (\textit{motion intention prediction}). Furthermore, we introduce two novel confidence-based criteria, optimized during the training phase, to enable dynamic stopping of the evidence accumulation, independently from movement speed or distance (\textit{motion direction prediction}). The two predictors run in parallel to provide external inputs to a \ac{fsm} that controls the robot according to operator's action, inaction, or motion direction.
Even if the intention and direction levels of prediction could be potentially included in the same predictor, as in other approaches~\cite{nikolaidis2013human}, we decided to implement them as separated processes and merge their predictions afterwards in the robot controller in order to maximize the performance in their specific tasks.
Finally, the proposed system has been tested with a UR10 manipulator robot in a human-robot collaborative task, where the human operator moves towards the location in front of his workspace and the robot goes coherently to the correspondent location in its workspace. For safety reason, we decided to test the system in a condition where the operator and the robot do not share the same workspace. Nevertheless, the proposed framework is general, thus it can be applied in many Industry 4.0 applications, where operators and collaborative robots can share the same workspace.
 

The remains of the paper is structured as follows. After discussing related work in \sect\ref{sec:related}, in \sect\ref{sec:intention} and in \sect\ref{sec:direction} we explain respectively motion intention and motion direction prediction methods in details. \sect\ref{sec:experimental} describes the experimental setup and the robot control architecture to evaluate the performance in a human-robot collaborative application. \sect\ref{sec:results} presents the results achieved during this work, followed by a more detailed discussion in \sect\ref{sec:discussion}. Finally, \sect\ref{sec:conclusions} summarizes the work and proposes further extensions.

\section{Related Work}
\label{sec:related}
Our work contributes to the field of human motion prediction for manipulation robots in industrial facilities. Several works focused on reconstructing human motion trajectories during reaching tasks~\cite{faraway2007modelling}~\cite{koppula2016anticipating}. Most of these systems require the knowledge of the end-point of the motion before predicting the whole trajectory. However, the reliable prediction of the target of a human motion from the early portion of the movement is still a challenging problem and the performance in reconstructing the trajectories strongly depends on the confidence of target prediction~\cite{perez2015fast}. Mainprice and Berenson~\cite{mainprice2013human} proposed a manipulation planning framework to predict the motion target by means of a \ac{gmm} for human-robot collaboration. Once the target has been identified, \ac{gmr} is used to extract the best fitting motion. The algorithm has been trained with a library of motions built from real Kinect data. They achieved $92\%$ of correct target classification after having processed $80\%$ of the trajectory, while performance were low in the early stages of motion.
Recently, classification frameworks based on Probabilistic Flow Tubes~(PFT) and Bayesian inference~\cite{perez2015fast}\cite{dong2012learning} have been proposed to improve the speed of the prediction using human joints' angles from a Vicon motion capture system. However, reliable predictions required to process more than half of the trajectory.

The aforementioned approaches propose to observe and track the body by means of visual systems, for example by exploiting 3D camera networks, or markers attached to the body. This solution has the drawback of being sensitive to camera occlusions, light variations, and motion blur~\cite{pfister2014comparative}. \acp{imu} are probably the main alternative to cameras, they are effectively used to learn new behaviors~\cite{fitter2016using} and control robots in industrial setups~\cite{yang2016neural}. 
In many cases, a multi-modal approach can be used to enrich the information and overcome limitations of uni-modal systems~\cite{stival2018toward}. Many solutions propose the introduction of physiological signals, recorded directly from the human body. 
\ac{emg} signals have been rarely considered as a unique tool for motion prediction~\cite{vogel2011emg} due to their non-stationarity and sensitivity to muscular fatigue and stress~\cite{de1997use}. Nevertheless, \ac{emg} are quite popular for controlling exoskeletons or prosthesis~\cite{stival2016online} and they have proved to be a valuable source of information in cooperative tasks when used in conjunction with other measurement units~\cite{peternel2016towards}.

In our work, the system exploits the information registered through a pair of Myo armbands, by Thalmic Labs, that enables simultaneous acquisition of kinematic and muscle activity information, at a very affordable cost. This device has already shown good acceptability in healthcare environments~\cite{esfahlani2018rehabgame}. However, few attempts have been made to introduce it in industrial environments~\cite{yang2016neural}. In the following, we will take into consideration the Myo multi-modal interface in combination with confidence-based criteria for dynamic stopping for early predictions of human motion in cooperative industrial tasks, while measuring feasibility and efficiency of the proposed methods.



\section{Motion intention prediction}
\label{sec:intention}
Literature on motion prediction focuses on selecting the correct motion given a set of samples. Almost all solutions consider these samples to be part of the motion to be recognized, discarding data where the user is still. The capability to predict motion continuously in time requires to understand when the operator is moving or not in order to start the recognizing the motion. To this aim, we divided our approach in two threads: \textit{motion intention prediction} and \textit{motion direction prediction}.
The first thread of the proposed system involves the training of a \ac{hmm} to predict continuously when the operator starts and ends a reaching movement. Two states of the model have been identified. (I)~A $REST$ state, where the operator does not perform an overt reaching movement. (II)~A $MOTION$ state, while the operator is performing the reaching task. The model has been trained by using the \textit{Baum-Welch} algorithm~\cite{baum1970maximization}. At each time step $t$, the posterior probability of the $i^{th}$ state, with $i = 1,2$ (i.e. rest or motion), is computed as:
\begin{equation}
    \resizebox{.9\hsize}{!}{$p(y_i(t)|x(t)) = p(x(t)|y_i(t))\sum_{j=1}^{2} p(y_i(t)|y_j)p(y_j(t-1))$}
\end{equation}
The overall filtered velocity magnitude has been exploited as observed variable $x(t)$ emitted by the state $y(t)$ at time $t$. This feature is obtained by computing the magnitudes of the angular velocity of the \ac{imu} sensors on the arm and forearm of the operator, and applying a zero-lag fourth-order band-pass filter between $0.01 Hz$ and $3 Hz$. Finally, the two filtered angular velocities are summed together. This metric is useful to reliably identify the presence of a generic motion involving the whole arm regardless of the specific motion that is performed. We decided not to use the \ac{emg} signal in this thread since different values of muscle activity could be present in both arm and forearm even if the subject is not moving the arm towards a target (i.e. isometric contractions during operations performed in the same location).

\section{Motion direction prediction}
\label{sec:direction}
\subsection{Feature selection}
The data available for prediction consist of inertial and muscular information. Each Myo device provides 3-axis angular velocity, 3-axis linear accelerometer and 8 built-in surface \ac{emg} electrodes, resulting in a total of $M=28$ features. In~\cite{perez2015fast} the authors used \ac{pca} for features selection. However, they also suggested that \ac{pca}-based feature selection alone is not guarantee to work in presence of significant nonlinearity within the data~\cite{wang2008gaussian}. Thus, we compared a number of dimensionality reduction techniques in order to limit the system complexity. 
\subsubsection{\acs{pca}}
\acf{pca} is based on an orthogonal linear transformation of the data in a different coordinate system that maximizes the uncorrelation between the new set of features, named principal components. The lower embedding dimension $C$ is chosen as the minimum number of components necessary to explain $90\%$ of the dataset variance. In this paper, \ac{pca} has been applied to all the 28 features from \ac{imu} and \ac{emg} channels (`\ac{pca}').
\subsubsection{\acs{nmf}}
\acf{nmf}~\cite{lee1999learning} algorithm has been extensively used in applications with dataset including only non-negative values, such as images and muscle activity envelopes.
Given a dataset $\textbf{X}$ of non-negative values, \ac{nmf} extracts $H$ and $W$ by minimizing the divergence $D($\textbf{X}$||HW)$ between the original and the reconstructed datasets. $H$ is the subject-specific synergy matrix and contains $C$ time-invariant and task-independent synergy modules. $W$ is the $C$-dimensional matrix of activation coefficients over time. The embedding dimension $C$ has been chosen by looking to the \acl{vaf}~\cite{d2003combinations} in order to have a robust agreement between the original and the reconstructed dataset. Due to its requisites, \ac{nmf} has been applied on the 16 \ac{emg} features (non-negative) concatenated to the features extracted by \ac{pca} on \ac{imu}~(possibly negative)~(`\ac{pca}\ac{nmf}').
\subsubsection{\acs{fda}}
Here, we refer to \acf{fda} not as the classification method, but as the dimensionality reduction method based on the Fischer's score. The aim of \ac{fda} is to project the data samples in a subspace with embedding dimension $C$ where the within-class variance is minimized, while the between-class variance is maximized, in order to improve class separability. 
Given a multi-class problem of $L$ classes, linear mapping matrix $W$ can be extracted from the first $C$ eigenvectors $\textbf{v}_i$ solving the system of linear equations:
\begin{align}
    S_W^{-1}\bar{S}\mathbf{v}_i = \lambda_i\mathbf{v}_i, \quad i=1,...,m \\
    \bar{S} = S_W + S_B
\end{align}
where $S_W$ and $S_B$ are the \textit{with-class scatter matrix} and the \textit{between-class scatter matrix}, respectively and $\lambda_i$ are descending eigenvalues. This solution is limited to $C<L$ cases, since the rank of $S_B$ is $L-1$, thus all the eigenvalues from $L$ to $M$ are the same and equal to 1~\cite{kung2014kernel}. In this paper, $C=L-1$ has been chosen and applied on two different sets of signals: the one including all the 28 features from \ac{imu} and \ac{emg} channels (`\ac{fda}'), and the one including only the 12 \ac{imu} features (`\ac{fda}-\ac{imu}').
\subsection{Gaussian Mixture Model and evidence accumulation}
Data from the feature selection phase have been used to train a probabilistic model, namely a \acf{gmm}, to estimate the direction chosen by the user by classifying among all the possible class of movements.
Given $N$ the total number of samples in the training dataset, a single data in input at the framework $\zeta_{n}$ with $1 \le n \le N$ can be written as:
\begin{equation}
\begin{split}
\zeta_n	&= \{\xi(n), \gamma_n \} \in \mathbb{R}^D, \ \ \  \xi(n) = \{\xi_{c} (n)\}^{C}_{c=1}
\end{split}
\label{eq:data}
\end{equation}
where $C = |\xi|$ is the number of selected features, $\xi(n) \in \mathbb{R}^C$ is the set of values assumed from the considered features, and $\gamma_n$ is the class associated to the sample.
$D = C + 1$ is the dimensionality of the problem.

The \ac{gmm} is completely represented by three parameters, i.e. mean, covariance and priors of each Gaussian component. The parameters are continuously optimized by means of the \ac{em} algorithm~\cite{dempster1977maximum}, seeded by K-means clustering.
The resulting \ac{pdf} is computed as:
\begin{equation}
p \left(\zeta_n \right) = \sum_{k=1}^{K}{\pi_k \, \mathcal{N} \left(\zeta_n; \mu_k, \Sigma_k \right)}
\label{eq:gmm}
\end{equation}
where $\pi_k$ are the priors probabilities, $\mathcal{N} \left(\zeta_n; \mu_k, \Sigma_k \right)$ represents the Gaussian distribution, and $K$ is the empirically estimated number of Gaussian components.
For each component, $\mu_k$ is the mean vector, and $\Sigma_k$ is the covariance matrix of the $k$-th distribution.

Samples belonging to a reaching movement are identified through the \textit{motion intention prediction} thread. They can be denoted as $\xi_0^{t} = \{\xi_0, \xi_1, ..., \xi_t\}$, going from movement onset to time instant $t$.
For direction classification purposes, we compute in real-time for each new sample $t$ the \ac{pdf} for each possible direction $l$ as 
$\rho_{t,l} = \text{PDF}(\xi_{t} | \gamma_{l})$,
where $ 0 \le t \le S$ is the index of the considered sample and $S$ is the total number of samples for a single reaching movement. $S$ depends on movement length and speed and can vary from movement to movement.
$\gamma_{l}$ with $1 \le l \le L$ indicates the index of all the possible $L$ directions.
For each sample, the \ac{pdf} is normalized $\forall l$, obtaining $\widetilde{\rho}_{t,l}$, so that $\sum_{l=1}^L \widetilde{\rho}_{l} = 1$.
Classification evidence is accumulated over time from movement onset up to time instant $t$ for each class~$l$ in order to improve the confidence on direction prediction:
\begin{equation}
\begin{split}
\alpha_{t,l} = \widetilde{\rho}_{t,l} + \sum\limits_{\tau=0}^{t-1} \widetilde{\rho}_{\tau,l}, \quad \forall l, \quad 1 \le i \le L.
\end{split}
\label{eq:sum}
\end{equation}
For each sample, the accumulated \ac{pdf} is normalized $\forall l$, obtaining $\widetilde{\alpha}_{t,l}$, so that $\sum_{l=1}^L \widetilde{\alpha}_{l} = 1$.
At each time instant, the chosen movement direction would be the one with the higher normalized accumulated probability:
\begin{equation}
\phi_{t} = l : max \left\lbrace \widetilde{\alpha}_{t,l}, \quad \forall l, \quad 1 \le l \le L \right\rbrace.
\label{eq:pdf_max_sum}
\end{equation}

\subsection{Criteria for dynamic stopping}
Following \eq\ref{eq:pdf_max_sum}, the predictions of \textit{motion direction prediction}, and thus the performance, depend on the time instant $t$ at which the evidence accumulation is stopped and the classification is performed. On one hand, the classifier should guarantee a certain accuracy in predicting the motion direction. On the other hand, the minimum accumulation time is difficult to determine \textit{a priori} since the movement is performed at self-selected speed. For this reason, two confidence-based criteria have been introduced to determine dynamically the time of accumulation for each movement. Given the vector of normalized accumulated posterior probabilities $\boldsymbol{\widetilde{\alpha}}_t = \{\widetilde{\alpha}_{t,1}, \widetilde{\alpha}_{t,2}, ..., \widetilde{\alpha}_{t,L}\}$ for each of the $L$ classes at time instant $t$, the first criterion, namely the \textit{ratio criterion}, is defined as:
\begin{equation}
    C_r(t) = 1 - \frac{k_2(t)}{k_1(t)}
\end{equation}
where
\begin{align}
    k_1(t) = \widetilde{\alpha}_{t,l_1} \quad with \quad l_1 = \underset{l \in L}{\arg\max} \; \widetilde{\alpha}_{t,l} \\
    k_2(t) = \widetilde{\alpha}_{t,l_2} \quad with \quad l_2 = \underset{l \neq l_1 \in L}{\arg\max} \; \widetilde{\alpha}_{t,l}.
\end{align}
It represents the ratio between the probabilities of the two most probable directions. This criterion has been introduced so that the system sends a command to the robot only if the confidence on the correspondent direction is sufficiently high. 
Given the cumulative sum of the raw posterior probabilities of the \ac{gmm} for each of the $L$ classes, which are the not normalized accumulated probabilities $\boldsymbol{\alpha}_t = \{\alpha_{t,1}, \alpha_{t,2}, ..., \alpha_{t,L}\}$, the second criterion, namely \textit{sum criterion}, is defined as:
\begin{equation}
    C_s(t) = \sum_{l=1}^{L}{\alpha_{t,l}}
\end{equation}
Combining the two criteria allows us to build the decision rule $T(C_r,C_s,th_r,th_s)$ as
\begin{equation}
    T(C_r,C_s,th_r,th_s) = \begin{cases} true, & \mbox{if } C_r > th_r \lor C_s > th_s \\ false, & \mbox{otherwise} \end{cases}
\end{equation}
The movement direction is predicted and sent to the robot as soon as one of the criteria is verified (i.e. $T=true$). A grid search has been conducted with a stratified 5-fold cross-validation for each subject to determine the thresholds $th_r$ and $th_s$. 

\section{Experiments and Robot control}
\label{sec:experimental}
\begin{figure}[t]
\centering
\includegraphics[width=0.9\columnwidth]{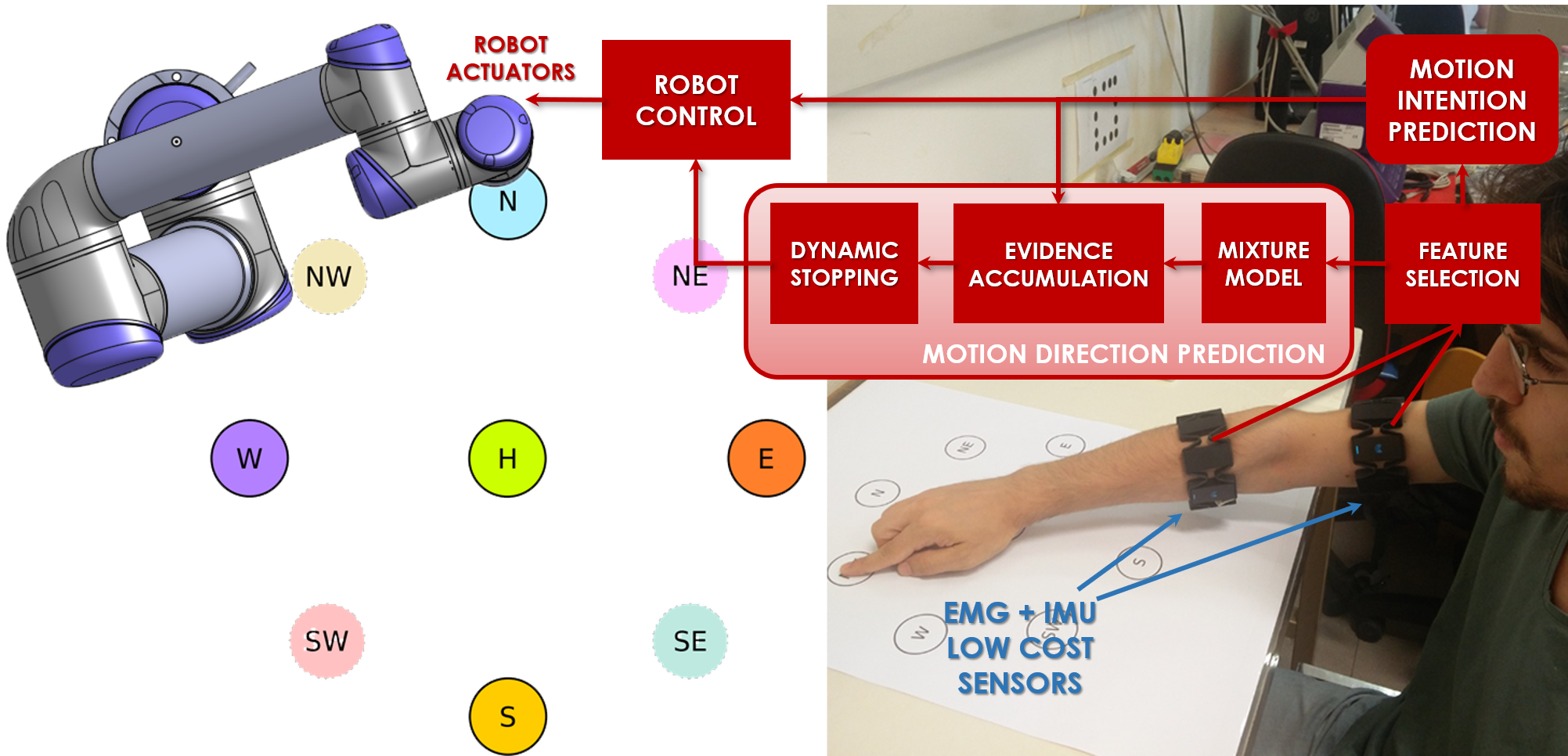}
  \caption{Experimental setup for performance evaluation. The subject is asked to move the hand towards one of the locations on a working bench. Once the human motion has been predicted, the robot goes coherently in the correspondent location. For these tests, the operator and the robot do not share the same workspace for safety reasons.}
\label{fig:experiment}
\end{figure}
\subsection{Experimental setup}
We performed a series of tests of the proposed framework on multiple subjects with the UR10 collaborative robot. 
Training data for models has been registered from four healthy right-handed subjects (age $26\pm4$, one female) performing reaching movements on a working bench with their right upper-limb. 
At each session, the subject is asked to move his hand from home position ('H') in the middle of the workspace, to one of the targets placed at a distance of $15~cm$ in cardinal directions ('N','E','S','W'). 
Each trial consists of a movement towards the target, followed by about 2 seconds of rest on the target, and a movement backwards the home position.
These main directions have been primarily chosen to compare the results of this study with a previous work~\cite{tortora2018synergy}. In addition to the cardinal directions, four secondary directions ('NE','SE','SW','NW') have been registered to test the robustness of the proposed system when increasing the number of directions to detect. For each subject, four sessions have been performed for training, resulting in 20 repetitions for each of the eight movements. Once the models have been trained, two sessions of testing for each subject have been performed to evaluate the performance of the system in real-time while controlling the UR10 robot~(\fig\ref{fig:experiment}). 

Angular velocity, linear acceleration and muscle activity information have been registered with two Myo armbands, from Thalmic Labs, worn on the upper arm and on the forearm. Both Myo devices have been connected and synchronized to the \ac{ros} middleware by means of a custom software library developed by our research group enabling the use of more than one device on the same PC\footnote{The library code is publicly available at~\url{https://github.com/ste92uo/ROS_multipleMyo}}.

\subsection{Movements segmentation}
Data collected from subjects during the training phase of the experimental protocol are composed of several trials acquired one after the other. 
Therefore, they present an alternation of states: motion (i.e. forward and backward movements) and rest (i.e. on the home position and on each target). We segmented the training dataset offline in order to distinguish among the two states and divide each sample accordingly. The segmentation consisted of two cascading steps, both composed by a filtering and a thresholding. The procedure has been applied to the sum of the angular velocity magnitude of both arm and forearm. The filters are respectively a zero-phase fourth-order Butterworth filter between $0.01 Hz$ and $3 Hz$, and a moving average filter with 1 second sliding window. The first threshold is computed as the average value plus six times the standard deviation of the first and last seconds of each session, where subjects are known to be in rest. The second threshold is fixed to $0.1$.

\subsection{Finite State Machine for robot control}
\begin{figure}
\centering
\includegraphics[width=0.8\columnwidth]{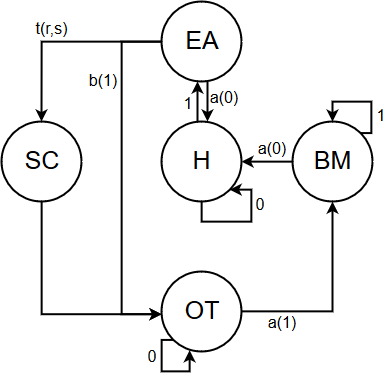}
  \caption{FSM Transition map. The transition values 0 and 1 corresponds to $REST$ and $MOTION$ respectively. Transition functions $a$ and $b$ enables changing the state after a predefined number of classifications of 0 and 1. $t(r,s)$ enables a state transition only if the criteria for dynamic stopping are satisfied.}
\label{fig:fsm}
\end{figure}
Once the movement intention and direction have been detected by the proposed prediction framework, the robot control depends on the specific scenario of applications. For our experiments, the robot is asked to continuously mirror the target locations where the operator moves the hand. Thus, the output of the \textit{motion intention prediction} and of the \textit{motion direction prediction} have been used to trigger a \acl{fsm} with the following states:
\begin{enumerate}
    \item \textit{Home (H)}: the subject is not moving from home position. The \textit{motion intention prediction} is predicting $REST$, thus the robot keeps its state;
    \item \textit{Evidence Accumulation (EA)}: the subject is moving towards a target. The \textit{motion intention prediction} is predicting $MOTION$ and the output probabilities of the \textit{motion direction prediction} are accumulated over time;
    \item \textit{Send Command (SC)}: the stopping criteria of the \textit{motion direction prediction} are met and the predicted movement direction is sent to the robot to activate the consequent action;
    \item \textit{On Target (OT)}: the subject is on the target position. The \textit{motion intention prediction} is predicting $REST$, thus the robot keeps its state;
    \item \textit{Back Movement (BM)}: the subject is moving back to the home position. The \textit{motion intention prediction} is predicting $MOTION$ and the robot goes back to its home position.
\end{enumerate}

\begin{figure}[t]
\centering
\includegraphics[width=0.9\columnwidth]{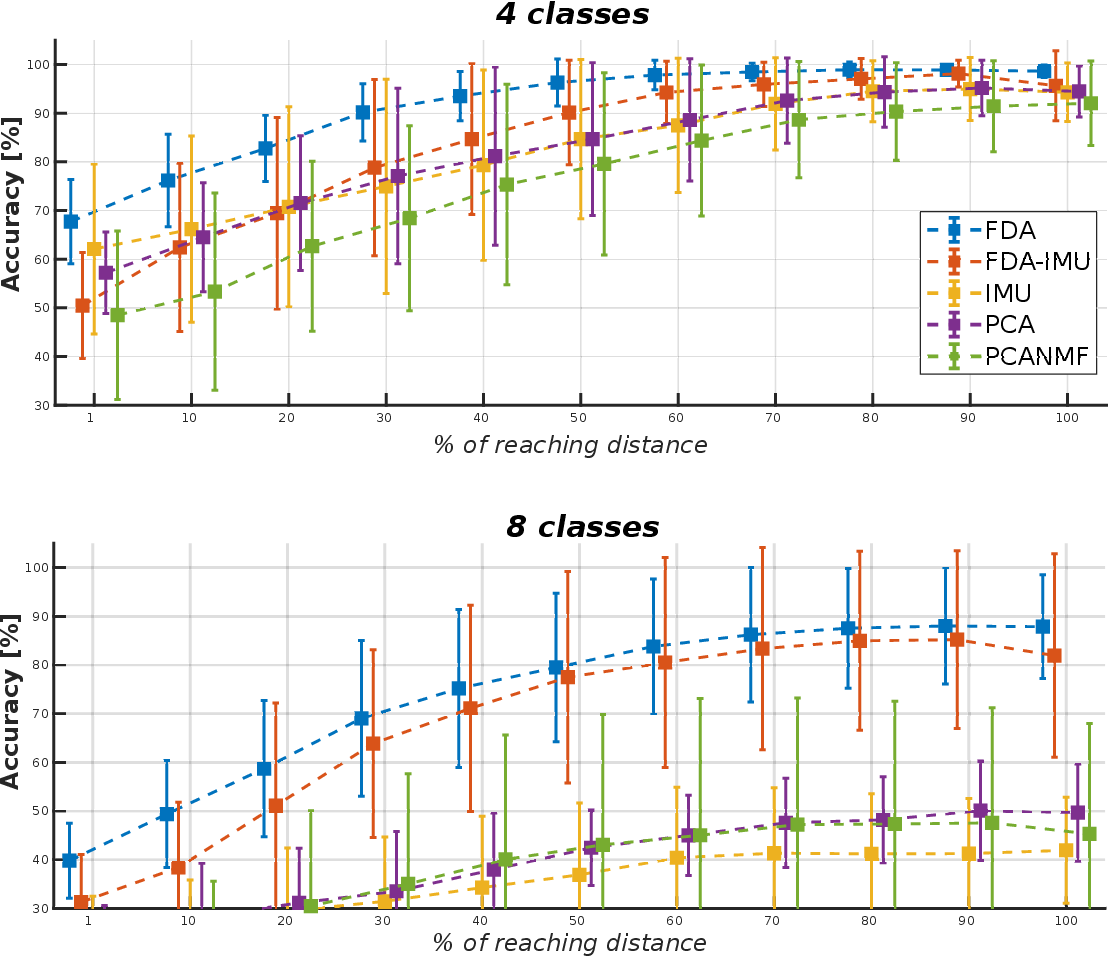}
  \caption{Classification accuracy over percentage of reaching distance of Gaussian Mixture Model coupled to 5 dimensionality reduction algorithms. The classifier has been tested with 4 classes (top) and 8 classes (bottom).}
\label{fig:accuracy}
\end{figure}
\begin{figure*}[t]
\centering
\includegraphics[width=0.8\textwidth]{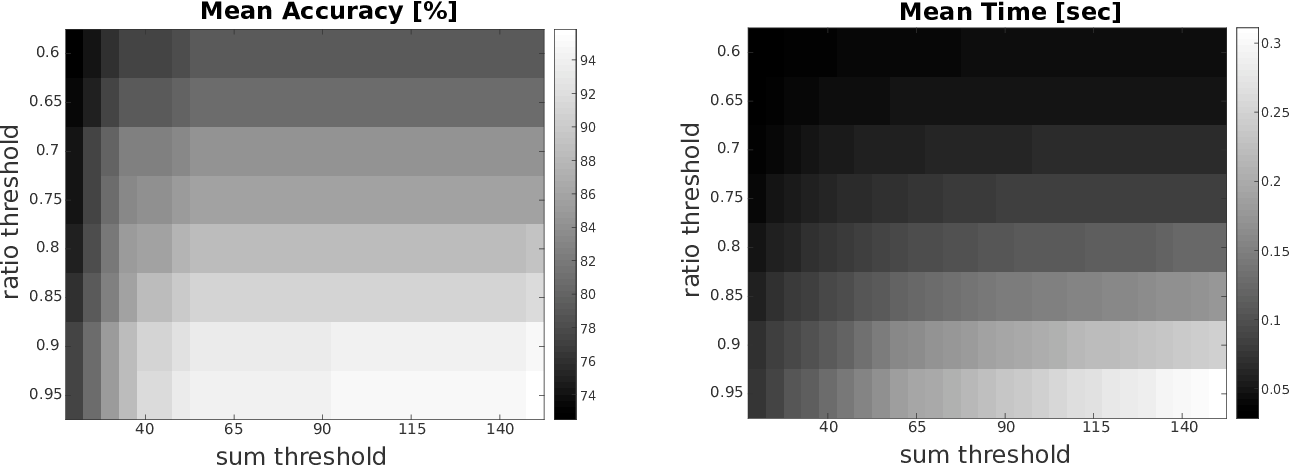}
  \caption{Visualization of the grid search results using the combined transition rule $T$. Each point of the picture reflects the performance that would have been achieved with the corresponding thresholds. Performance have been measured in terms of mean accuracy (left) and mean time to send a command (right).}
\label{fig:gridsearch}
\end{figure*}
The transition map of the proposed \ac{fsm} is shown in \fig\ref{fig:fsm}. At the beginning of each session, the subject is supposed to be in \textit{Home} state and the \ac{hmm} is initialized to $REST$. When $MOTION$ is detected by the \ac{hmm}, the machine immediately transits to \textit{Evidence Accumulation} state. If the accumulated evidence fulfill the criteria for dynamic stopping ($t(r,s)$), the machine goes to \textit{Send Command} state. On the other hand, if the \ac{hmm} identifies a $REST$ and it is kept for at least $X$ samples ($a(0)$), the machine goes back to \textit{Home} state. 
If the accumulated evidence does not satisfy the criteria for longer than $Y$ samples ($b(1)$), the \ac{fsm} jumps directly to \textit{On Target} state, meaning that the subject moved towards a target but the direction has not been identified with enough confidence, thus no command is sent to the robot.
The state goes to \textit{On Target} from \textit{Send Command} state immediately after the predicted direction is sent to the robot controller. If the machine is in \textit{On Target} state and the \ac{hmm} detects a $MOTION$ for at least $X$ samples ($a(1)$), state transits to \textit{Back Movement}. It stays in \textit{Back Movement} until going back \textit{Home} once \ac{hmm} detects a $REST$ for at least $X$ samples ($a(0)$).
The values of $X$ and $Y$ have been manually tuned for each subject as a fraction of the average number of $MOTION$ samples within a motion state and the number of $REST$ samples within a rest state.
\begin{table}[h]
\caption{Selected thresholds and corresponding performance for each subject in the 4 classes case.}
\label{tab:gridsearch}
\centering
\begin{adjustbox}{max width=\textwidth}
\begin{tabular}{l | ccccc}
 \textbf{Sbj.} & \boldmath{$th_r$} & \boldmath{$th_s$} & \textbf{acc. [\%]} & \textbf{time [sec]} & \textbf{perc.}\\
 \hline \hline
\rule{0pt}{3ex}
s1 & 0.95 & 95 & 95.0 $\pm$ 4.8 & 0.24 $\pm$ 0.15 & 20\% \\
\hline
\rule{0pt}{3ex}
s2 & 0.95 & 35 & 95.0 $\pm$ 1.8 & 0.37 $\pm$ 0.24 & 32\% \\
\hline
\rule{0pt}{3ex}
s3 & 0.95 & 45 & 95.0 $\pm$ 3.5 & 0.25 $\pm$ 0.13 & 22\% \\
\hline
\rule{0pt}{3ex}
s4 & 0.95 & 25 & 95.0 $\pm$ 3.5 & 0.16 $\pm$ 0.17 & 14\% \\
\end{tabular}
\end{adjustbox}
\end{table}

\section{Results}
\label{sec:results}
\begin{table*}[h]
\caption{Summary of the comparison between the proposed method and relevant state-of-art methods.
Abbreviations: $N_{dem}$ (number of demonstrations per class), $N_{cl}$ (number of classes), proc.time (processing time per time step), req.conf. (requested prediction confidence), acc. (prediction accuracy), corr.class. (percentage of correct classifications), perc.traj. (percentage of processed trajectory).}
\label{tab:comparison}
\centering
\begin{adjustbox}{max width=\textwidth}
\begin{tabular}{l | ccccc}
 \textbf{Method} & \boldmath{$N_{dem}$} & \boldmath{$N_{cl}$} & \textbf{proc.time[ms]} & \textbf{req.conf.: acc.[\%]/time[ms]} & \textbf{corr.class.[\%]/perc.traj.[\%]}\\
 \hline \hline
\rule{0pt}{2ex}
                                      &    &   &   &   & 50/43 \\
\ac{gmm} in~\cite{mainprice2013human} & 24 & 8 & - & - & 80/60 \\
                                      &    &   &   &   & 92/80 \\
\hline
\rule{0pt}{2ex}
                            &    &   &      &          &  80.00/16.60\\
PFT in~\cite{perez2015fast} & 13 & 4 & 5.01 & 80\%: 70/416.6 &  90.12/54.44\\
\hline
\rule{0pt}{2ex}
                            &    &   &      &   &  71.21/43.64\\
PFT in~\cite{perez2015fast} & 10 & 9 & 6.03 & - &  79.09/60.00\\
                            &    &   &      &   &  90.00/80.00\\
\hline
\rule{0pt}{2ex}
          &    &   &      & 85\%: 85.2/93.2  & 83/20 \\
FDA + GMM & 20 & 4 & 2.53 & 90\%: 90.2/136.4 & 94/40 \\
          &    &   &      & 95\%: 95.0/255.1 & 98/60 \\
\hline
\rule{0pt}{2ex}
          &    &   &      & 80\%: 80.5/382.2 & 70/30 \\
FDA + GMM & 20 & 8 & 2.89 & 85\%: 86.4/570.1 & 85/60 \\
          &    &   &      & 90\%: 89.9/708.9 & 90/80 \\
\end{tabular}
\end{adjustbox}
\end{table*}
\subsection{Offline performance evaluation}
To assess the performances of the system, the classification accuracy computed over the reaching distance has been computed by 5-fold cross-validation and averaged across subjects. The performance in discriminating between the four main directions for the different feature selection methods, all coupled with the \ac{gmm}, are shown in \fig\ref{fig:accuracy}~(top). As expected, the accuracy increases over time, reaching more than $90\%$ of accuracy for all the methods after processing $80\%$ of the movement. However, `FDA' shows the highest performance in the first half of the reaching distance when compared to the other methods. The accuracy for `FDA' reached more than $90\%$ of accuracy already at $30\%$ of reaching distance, and with a remarkably lower variability between subjects. To assess robustness and scalability of the classification methods, we tested their performance over the extended set of eight classes. The results, shown in \fig\ref{fig:accuracy}~(bottom), reveal a higher robustness with the number of classes of `FDA' and `FDA-IMU'. In fact, the accuracy is higher than $80\%$ at $50\%$ of reaching distance, and it is up to $90\%$ after processing $80\%$ of the movement. The highest performance have been achieved by 'FDA', thus the couple FDA-\ac{gmm} will be considered as the selected classifier for the following analysis.

The results of the grid search for the thresholds of stopping criteria are shown in \fig\ref{fig:gridsearch} for one of the tested subjects. Ideally, we would like to find the pair of thresholds that maximizes the accuracy while minimizing the time, thus maximizing the information transfer rate. It can be noticed that, at increasing of both thresholds, the accuracy increases, as well as the time to send the command. The selection of the thresholds enables a flexible design of the classifier's performance, adjusting the speed-accuracy trade-off according to the application. In this context, the couple of thresholds has been selected as the one that guarantees an average accuracy of at least $95\%$ in the minimum amount of time. The selected thresholds for each subject and their corresponding accuracy, time and percentage of trajectory performance are provided in \tab\ref{tab:gridsearch} for the four classes case.
\begin{figure}[h]
\centering
\includegraphics[width=\columnwidth]{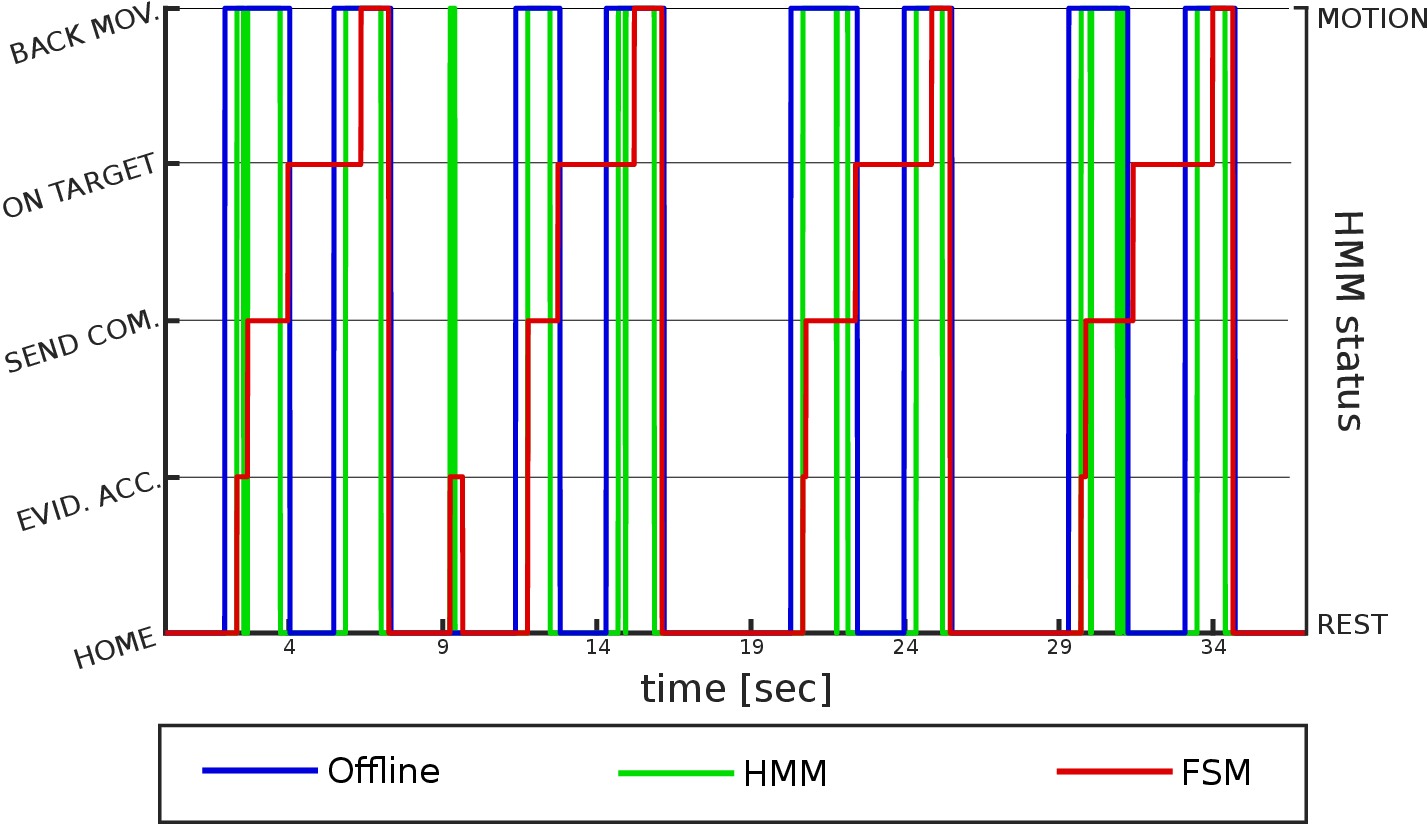}
  \caption{A sample of four trials taken from the final experiment with the \ac{fsm} used to control the UR10 robot. The offline segmentation (blue) is shown as ground truth for the evaluation. The trained \ac{hmm} (green) has been used to predict states of $MOTION$ and $REST$. The \ac{fsm} (red) correctly transits through its possible states according to the outputs of two motion prediction levels.}
\label{fig:iterations}
\end{figure}
\subsection{Real-time performance evaluation}
In the final testing, the two \ac{gmm} and \ac{hmm} classifiers have been used as external inputs to the \ac{fsm}, implemented on ROS to control a physical manipulator robot so that it reaches one among four locations in cardinal directions. The location in the robot workspace is predicted in real-time by the system and it should correspond to the location reached by the operator. 
The results of these testing are shown in \fig\ref{fig:iterations}. The figure shows four consecutive trials, that consists of four commands to control the robot. The blue line represents the offline movement segmentation, and it can be either $REST$ or $MOTION$. Two consecutive motions represent a forward movement towards one of the target, and a backward movement to the home position. The green line represents the output of the \ac{hmm}, that can predict either a status of $REST$ or a status of $MOTION$, resulting in an average accuracy of $82.5\pm4.8\%$ across subjects. 
The red line represents the transitions of the \ac{fsm} through its states (\textit{Home}, \textit{Evidence Accumulation}, \textit{Send Command}, \textit{On Target}, \textit{Back Movement}). It is possible to notice that the \ac{fsm} transits correctly through the states, driven by the \ac{gmm} and the \ac{hmm} and following the operator's movement. Quantitatively, the performance of the whole system in real-time averaged across the tested operators results in $94.3\%\pm2.9\%$ of accuracy in predicting motion direction. On average, the system requires $160msec\pm80msec$ to met the stopping criteria and send the command to the robot. In case of transition error, the \ac{fsm} has been manually reset to the \textit{Home} state and the experiment has been restarted from the following trial. The percentage of erroneous transitions of the \ac{fsm} during the whole experiment (i.e. 240 expected transitions) is $1.6\%\pm1.5\%$ on average for each operator. The average processing time per time step was $2.53msec$~($2.89msec$ with 8 classes), of which most of the time is taken by the mixture model classification, about $1.5msec$~($1.9msec$ with 8 classes). The Myo armbands provide a sampling frequency of $100 Hz$, requiring the system a processing time of $10msec$, thus the time performance of the system was less than $30\%$ of the time requirement. The tests have been conducted using an Intel Core i5-4210M (2.60 GHz x 4) and 8 GB of RAM.

\section{Discussion}
\label{sec:discussion}
Different feature selection algorithms have been evaluated offline to select the approach to be used in the \textit{motion direction level} of the prediction. In particular, the \ac{nmf} has been tested since it has previously shown to be efficient in extracting motion primitives from \ac{emg} envelopes and to improve classification accuracy and speed in a similar context~\cite{tortora2018synergy}. However, \ac{nmf} on \ac{emg} data, coupled to \ac{pca} on \ac{imu} data, performed poorer in this application compared to other methods. On the other hand, \ac{fda} has been tested with and without the contribution of \ac{emg} channels, to see if inertial information alone would be enough for motion classification. Interestingly, in both the 4 classes and 8 classes cases, \ac{fda} with additional information on muscle activity improved the classification accuracy of about $15\%$ and with lower variability across subjects. These findings strengthen the hypothesis that multi-modal approaches, enriched with the introduction of physiological signals, can overcome the limitations of traditional uni-modal approaches.

\tab\ref{tab:comparison} shows a comparison of our results with relevant state-of-art methods (1st, 2nd and 3rd rows). Our system based on \ac{fda}-\ac{gmm} with evidence accumulation provides a higher percentage of correctly classified movements considering different fixed time limits for the predictor (last column). The evidence accumulation is a well-known solution to improve the accuracy of classification systems, particularly useful in applications where the driving signals are very noisy~\cite{lenhardt2008adaptive}. However, it could be challenging to determine a time to stop the accumulation \textit{a priori}. In~\cite{perez2015fast}, the time limit for the prediction was fixed to $416.6msec$ and determined from the training dataset to achieve an accuracy of $80\%$ at least, while the performance during experiments achieved only $70\%$ of accuracy. The proposed confidence-based criteria enables a dynamic stopping of the predictor in order to adjust the trade-off between early-anticipation and accuracy according to the application~(5th column). Thanks to this approach, the time limit for the predictor is identified separately for each movement. The accuracy is $95.0\%$ after $255.0msec$ on average for the four classes case, which is a little more than $22\%$ of the trajectory from movement onset. Similar performance have been found in the real-time testing with the physical robot.
Different accuracy and time performance can be achieved by changing the requested prediction confidence (i.e. 85\%, 90\%, 95\% for the four classes case, 80\%, 85\%, 90\% for the eight classes case).
Results outperform the current state-of-art methods. 
It is worth to notice the filtering effect of the \ac{fsm} on the \ac{hmm} misclassification, discarding fast and unstable transitions from $REST$ to $MOTION$ and viceversa, according to the machine state. This can be seen clearly between the first and second trials of \fig\ref{fig:iterations}, where a $MOTION$ peak of the \ac{hmm} activates the \textit{Evidence Accumulation} state immediately, but it does not last enough to allow the \ac{gmm} to verify the stopping criteria. As a consequence, the machine goes back to the \textit{Home} state, without sending undesired commands to the robot.

\section{Conclusions}
\label{sec:conclusions}
In this paper, we presented a human-machine interface to predict human motion in industrial applications. The interface architecture is based on two separated predictors for movement intention and movement direction. As proof-of-concept, the interface has been used to trigger a \ac{fsm} to control a UR10 collaborative robot. 
The system has been implemented under ROS for an easier applicability to several robotic devices and applications. The main contribution of our approach is the flexibility in the trade-off between early-anticipation and accuracy thanks to the novel confidence-based criteria for dynamic stopping.
In future works, we plan to improve the system by making models parameters and stopping thresholds able to generalize among multiple subjects~\cite{stival2018subject}, to avoid long and costly training sessions for each operator.


\bibliographystyle{IEEEtran}
\bibliography{my_biblio}

\end{document}